\documentclass[%
 aip,
 amsmath,amssymb,
 reprint,%
]{revtex4-1}

\usepackage{graphicx}
\usepackage{dcolumn}
\usepackage{bm}

\usepackage{mathptmx}
\usepackage{etoolbox}

\makeatletter
\def\@email#1#2{%
 \endgroup
 \patchcmd{\titleblock@produce}
  {\frontmatter@RRAPformat}
  {\frontmatter@RRAPformat{\produce@RRAP{*#1\href{mailto:#2}{#2}}}\frontmatter@RRAPformat}
  {}{}
}%
\makeatother
\begin{document}

\preprint{AIP/123-QED}

\title[Internal noise in spiking neural networks]{General aspects of internal noise in spiking neural networks}
\author{I.D. Kolesnikov}%
\author{D.A. Maksimov}%
\author{V.M. Moskvitin}%
\author{N. Semenova}%
 \email{semenovani@sgu.ru}
 \affiliation{Saratov State University, Astrakhanskaya str. 83, Saratov 410012, Russia}%

\date{\today}

\begin{abstract}
This study examines the impact of additive and multiplicative noise on both a single leaky integrate-and-fire (LIF) neuron and a trained spiking neural network (SNN). Noise was introduced at different stages of neural processing, including the input current, membrane potential, and output spike generation. The results show that multiplicative noise applied to the membrane potential has the most detrimental effect on network performance, leading to a significant degradation in accuracy. This is primarily due to its tendency to suppress membrane potentials toward large negative values, effectively silencing neuronal activity. To address this issue, input pre-filtering strategies were evaluated, with a sigmoid-based filter demonstrating the best performance by shifting inputs to a strictly positive range. Under these conditions, additive noise in the input current becomes the dominant source of performance degradation, while other noise configurations reduce accuracy by no more than 1\%, even at high noise intensity. Additionally, the study compares the effects of common and uncommon noise across neuron populations in hidden layer, revealing that SNNs exhibit greater robustness to common noise. Overall, the findings identify the most critical noise mechanisms affecting SNNs and provide practical approaches for improving their robustness.
\end{abstract}

\maketitle

\begin{quotation}
As artificial neural networks and their tasks grow increasingly, complex computing systems may soon approach a scalability limit where existing hardware cannot meet rising demands. Maximum efficiency can be achieved through fully hardware-implemented neural networks, where neurons correspond to physical nonlinear components and connections to communication channels, eliminating memory access and large-scale numerical operations. However, such implementations fundamentally change noise behavior: unlike digital systems, where noise mainly enters through inputs, hardware neural networks are affected by multiple internal noise sources with diverse properties.

Studying different types of noise in machine learning is therefore essential to understand which disturbances are amplified or suppressed in artificial neural networks, and how they influence learning and inference. This is particularly important for developing robust neural systems capable of operating under stochastic perturbations, especially in real-world hardware prototypes that are inherently subject to both internal and external noise.
\end{quotation}

\section{Introduction}\label{sec:intro}
As artificial neural networks (ANNs) and their tasks become increasingly complex, we may approach a point where modern computational resources are insufficient to meet growing demands \cite{Hasler2013,Gupta2015}. A promising alternative is the emerging paradigm of hardware neural networks \cite{Karniadakis2021}, in which networks are implemented as physical systems rather than software. In this approach, neurons and their connections are realized directly at the physical level, governed by underlying physical principles. Currently, the most efficient implementations are based on laser systems \cite{Brunner2013a}, memristors \cite{Tuma2016}, and spin-transfer oscillators \cite{Torrejon2017}.

The physical implementation of artificial neural networks (ANNs) in real experimental systems is a relatively recent direction in machine learning \cite{Psaltis1990,Tuma2016,Torrejon2017,Bueno2018}. Such systems offer significant advantages in computational speed \cite{Chen2023,Aguirre2024}, but may also exhibit intrinsic noise arising from components of the experimental setup. This noise can propagate through the network and potentially accumulate, making the accuracy of such systems worse. Importantly, this type of noise should not be confused with noise present only in input data \cite{Shen2017,Tait2017}. While several studies report that internal noise degrades performance \cite{Frye1991,Dibazar2006,Soriano2015,Moon2019,Janke2020}, our recent work \cite{Kolesnikov2026CSF} demonstrates that, in some cases, training deep and recurrent networks with internal noise can improve their robustness to noise in future.

Although the structure of ANNs was originally inspired by a simplified model of the brain, they differ fundamentally from biological neural networks in architecture, neuronal dynamics, and learning rules. This has led to the development of spiking neural networks (SNNs), often referred to as the third generation of neural networks, with the potential to overcome key limitations of conventional ANNs.
SNNs constitute a class of ANNs in which neurons communicate via sequences of spikes with varying timing and frequency \cite{GhoshDastidar2009,Ponulak2011}. From the perspective of energy efficiency, their implementation on neuromorphic hardware (such as TrueNorth \cite{Merolla2014}, Loihi \cite{Davies2018}, SpiNNaker \cite{Furber2014}, and NeuroGrid \cite{Benjamin2014a}) represents a promising approach \cite{Yamazaki2022}.

In this work, we investigate the impact of internal noise on spiking neural networks (SNNs). Both multiplicative and additive noise are considered, introduced in common and uncommon forms. The choice of these noise types is motivated by optical implementations of ANN described in Ref.~\cite{Bueno2018}. While our study is not limited to optical systems, we examine a broad range of noise intensities and their ratios, enabling the results to be generalized to hardware neural networks of various physical realizations.

\section{Single neuron}\label{sec:single}

A wide range of neuron models with varying dynamical exactness and complexity can be employed as elementary units of SNNs. Biophysically grounded models such as the Hodgkin-Huxley, Hindmarsh-Rose, FitzHugh-Nagumo, and Izhikevich systems provide the closest correspondence to real neurons. While these models capture biological behaviour more accurately, their high computational cost significantly slows training and testing.

Among models that provide a balance between computational efficiency, spiking behavior, and suitability as artificial neurons, the leaky integrate-and-fire (LIF) model is widely used. It accumulates weighted inputs similarly to a conventional artificial neuron. However, instead of applying an activation function directly, it integrates the input over time with leakage, analogous to an RC circuit. When the membrane potential exceeds a threshold, the neuron emits a spike. The LIF model abstracts away the spike waveform, treating it as a discrete event. Consequently, information is encoded in spike timing or rate rather than amplitude or signal's form. Such simplified spiking models have yielded key insights into neural coding, memory, network dynamics, and, more recently, deep learning. For these reasons, the LIF neuron is commonly adopted for training and evaluation of SNNs \cite{snntorch}.

\subsection{Model}\label{sec:single:model}

A variety of neuron models can be classified within the LIF framework. In this work, we consider the first-order LIF neuron. The model is defined by a set of equations relating the input current, the membrane potential, and the output spikes. The input current instantaneously contributes to the membrane potential:
\begin{equation}\label{eq:lif_cur_mem}
\begin{array}{c}
U_\mathrm{mem}[t+1] = \beta U_\mathrm{mem}[t] + (1-\beta) I_\mathrm{in}[t+1] ; \\
I_\mathrm{in}[t] = WX[t].
\end{array}
\end{equation}
Here, $\beta$ denotes the membrane potential decay rate, $\beta = U_{\mathrm{mem}}[t+1]/U_{\mathrm{mem}}[t]$. $I_\mathrm{in}[t]$ is the input current including the information about spikes and signal from previous layer. This can be interpreted in the following way. $X[t]$ is an input voltage, or spike, and is scaled by the synaptic conductance of $W$ to generate a current injection to the neuron. This gives the following result:
\begin{equation}\label{eq:lif_noreset}
U_\mathrm{mem}[t+1] = \beta U_\mathrm{mem}[t] + WX[t+1]= \beta U_\mathrm{mem}[t] + I_\mathrm{in}[t+1].
\end{equation}

We now describe spike generation and the membrane potential reset mechanism. A spike is emitted when the membrane potential exceeds a threshold $U_{\mathrm{threshold}}$:
\begin{equation}\label{eq:lif_spk}
S[t] = \Biggl\{
\begin{array}{l}
 1 \text{, if } U_\mathrm{mem}[t]>U_\mathrm{threshold} \\
 0 \text{, otherwise }
\end{array}
\end{equation}
Equation (\ref{eq:lif_noreset}) shows that without a reset mechanism, $U_{\mathrm{mem}}$ grows continuously, eventually causing the spike condition in (\ref{eq:lif_spk}) to be satisfied constantly, leading to continuous spiking. To prevent this uncontrolled growth, a membrane potential reset mechanism is introduced. The reset-by-subtraction mechanism is modeled by:
\begin{equation}\label{eq:lif_reset}
U_\mathrm{mem}[t+1] = \beta U_\mathrm{mem}[t] + I_\mathrm{in}[t+1] - S[t]U_\mathrm{threshold}.
\end{equation}

For network training and simulation, we use Python with the snnTorch library \cite{snntorch}. In this library, the model corresponds to the ``Leaky'' class with $\beta = 0.9$. Throughout this work, the threshold is also fixed at $U_{\mathrm{threshold}} = 1$.

\subsection{Internal noise in a single neuron}\label{sec:single:noise}
There are several ways to introduce internal noise into the neuron model. Noise can affect a spiking neuron at three stages. One common source is the input signal, analogous to interference in the neuron's input channel or noisy connections from upstream neurons. This type of noise is applied to the input current, $I_{\mathrm{in}}[t]$. In this work, we consider two forms of input noise: additive and multiplicative:
\begin{equation}\label{eq:noise_cur}
I^*_\mathrm{in}[t] = I_\mathrm{in}[t]\cdot(1+\sqrt{2D_M}\xi_M[t]) + \sqrt{2D_A}\xi_A[t].
\end{equation}
Additive noise $\sqrt{2D_A}\xi_A[t]$ is added to the signal, whereas multiplicative noise $\sqrt{2D_M}\xi_M[t]$ scales it. Here, $\xi_A[t]$ and $\xi_M[t]$ are zero-mean, unit-variance Gaussian white noise sources. The variance of the noise influence is controlled by the noise intensities $D_A$ and $D_M$.

Similarly, noise can be applied directly to the membrane potential, $U_\mathrm{mem}[t]$:
\begin{equation}\label{eq:noise_mem}
U^*_\mathrm{mem}[t] = U_\mathrm{mem}[t]\cdot(1+\sqrt{2D_M}\xi_M[t]) + \sqrt{2D_A}\xi_A[t].
\end{equation}

Noise can also be introduced in the spiking output signal $S[t]$:
\begin{equation}\label{eq:noise_spk}
S^*[t] = S[t]\cdot(1+\sqrt{2D_M}\xi_M[t]) + \sqrt{2D_A}\xi_A[t].
\end{equation}
Introducing noise at this stage has little effect for an isolated neuron, as it does not alter spike generation. However, in the next section, we revisit this aspect when examining the impact of noise on a network of interconnected neurons.

\subsection{Positive input signal}\label{sec:single:positive}
We begin by examining how noise in the input current (\ref{eq:noise_cur}) and membrane potential (\ref{eq:noise_mem}) affects the signal and behavior of a single neuron. For this purpose, we consider a periodic positive input signal:
\begin{equation}\label{eq:input_positive}
I_\mathrm{in}[t] = \frac{1}{2} \big(1+\sin[3\pi t]\big).
\end{equation}

\begin{figure}[t]
\includegraphics[width=\linewidth]{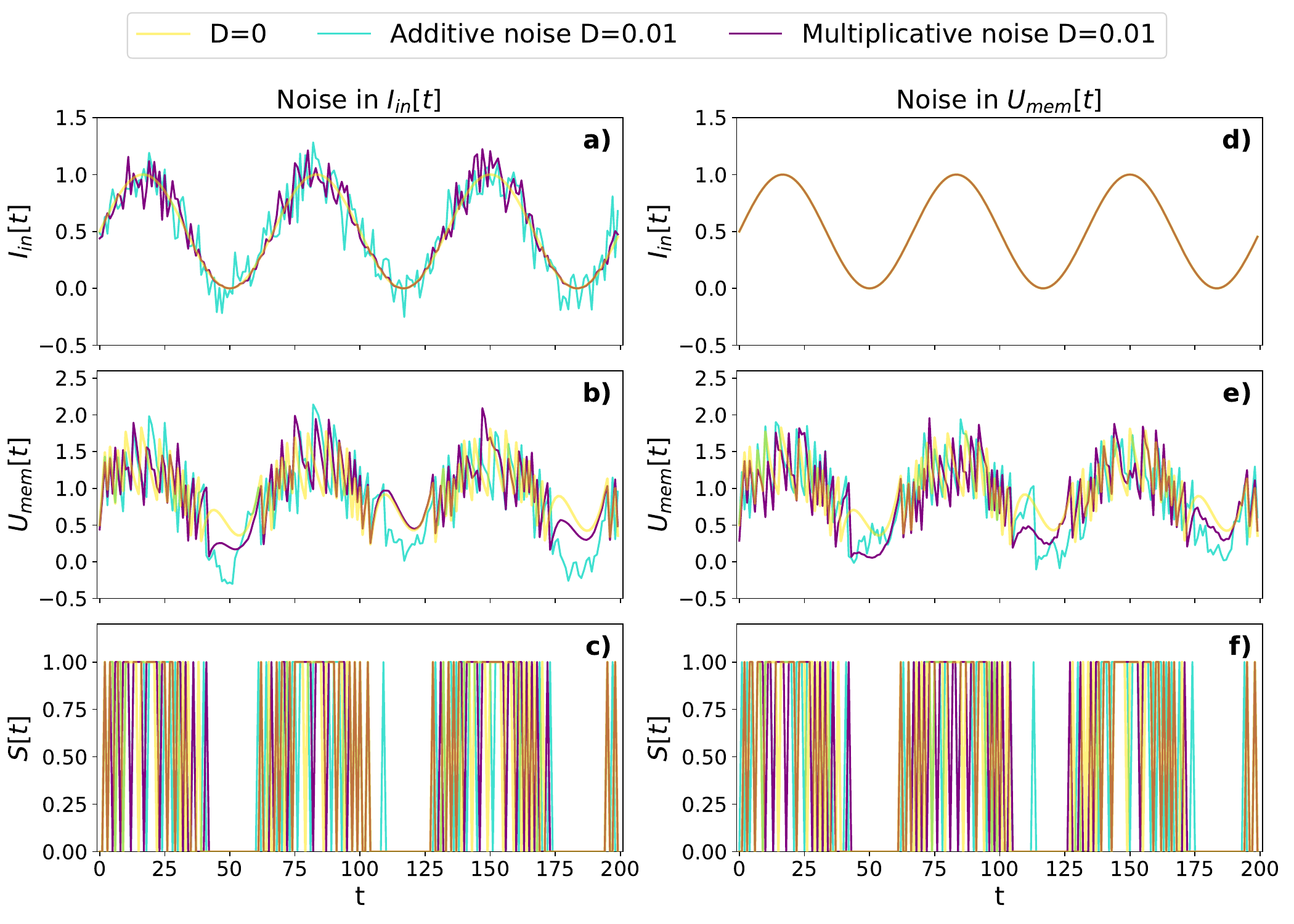}
\caption{\label{fig:single:positive:signal} Effect of noise applied to the input current (a–c) and membrane potential (d–f). Panels show the input current (a, d), membrane potential (b, e), and output spikes (c, f) under additive noise (cyan) and multiplicative noise (purple). The noise-free case is shown in yellow. The figure corresponds to a positive input current (\ref{eq:input_positive}).}
\end{figure}

The left panels of Fig.~\ref{fig:single:positive:signal} show the input current $I_\mathrm{in}[t]$ (a), the membrane potential $U_\mathrm{mem}[t]$ (b), and the output spikes $S[t]$ when the noise is introduced into the input current. Cyan lines correspond to additive noise with intensity $D_A = 10^{-2}$, while purple lines represent multiplicative noise. Yellow lines indicate the noise-free case. The right panels, Fig.~\ref{fig:single:positive:signal}(d–f), illustrate the same quantities when noise is applied to the membrane potential $U_\mathrm{mem}[t]$.

The plots show that additive noise introduces a persistent deviation throughout the signal, whereas multiplicative noise becomes more pronounced as the input approaches 1. When the input signal nears 0, the effect of multiplicative noise is almost negligible.

\begin{figure}[t]
\includegraphics[width=\linewidth]{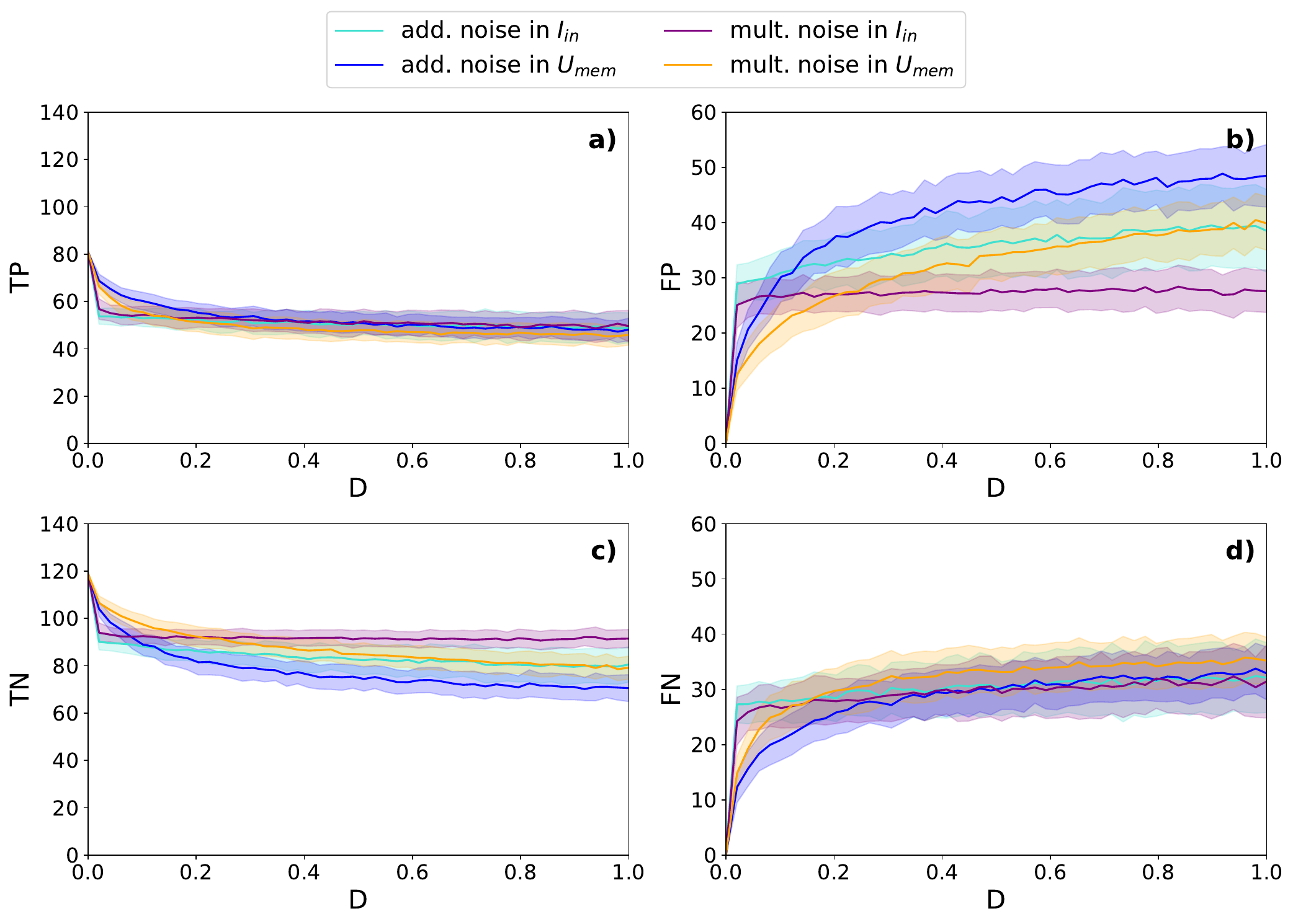}
\caption{\label{fig:single:positive:rate} True-positive (a), false-positive (b), true-negative (c) and false-negative (d) events for a single neuron in the case additive noise influence in input current (cyan curves) or membrane potential (blue lines); and multiplicative noise influence in input current (purple lines) or membrane potential (orange lines). The curves were obtained for a purely positive input signal (\ref{eq:input_positive}). }
\end{figure}

Regarding the output spikes, $S[t]$, Fig.~\ref{fig:single:positive:signal}(c,f) shows that even with noise intensity $D = 0.01$, spike failures occur. However, this representation does not allow for a quantitative assessment of the error. To address this, we use a classification-based approach with true positives (TP), true negatives (TN), false positives (FP), and false negatives (FN), taking the noise-free signal as the reference. A `positive' event corresponds to a spike, and `negative' to its absence. The resulting dependence of these metrics on noise intensity is shown in Fig.~\ref{fig:single:positive:rate}.

In Fig.~\ref{fig:single:positive:rate}, cyan curves correspond to additive noise applied to the input current, while blue curves represent additive noise in the membrane potential. Purple and orange curves show multiplicative noise applied to the input current and membrane potential, respectively.

Figure~\ref{fig:single:positive:rate} shows that many of the curves intersect, allowing the effects of noise to be roughly categorized into those observed at low and high noise intensities. On most panels (a, c, d), the metrics approach saturation. Focusing on the true-positive rate (TP, panel a), even at high noise intensities, TP decreases by roughly 25 units ($\sim 30\%$). False positives (FP, panel b) corresponding to the rate of error spikes are also significant. For multiplicative noise in $I_\mathrm{in}$, FP reaches nearly the same value of 25, suggesting that if coding relies on total spike count, multiplicative noise in $I_\mathrm{in}$ keeps the overall number of spikes on almost the same level. Regarding false spikes at low noise intensities, additive noise in $I_\mathrm{in}$ (cyan curve) is the most critical. At higher intensities, additive and multiplicative noise in $U_\mathrm{mem}$ (blue and orange curves) dominate the false-positive behavior.

\subsection{Positive/negative input signal}\label{sec:single:posneg}
In the previous section, we considered only a purely positive input signal. However, in SNNs, inputs can also be negative. For instance, in the NMNIST dataset, the handwritten MNIST digits are adapted for SNNs allowing both positive and negative spikes.

The main issue with negative inputs is that prolonged negative input currents can drive the neuron toward ``death.'' While spiking neurons typically include a reset mechanism to prevent continuous spiking, this mechanism does not cover negative potentials in LIF neurons. As a result, the membrane potential can drift to unbounded negative values. This behavior can be useful for training, but it complicates interpretation of noise effects, since the noise intensity becomes difficult to relate to the overall signal level.

\begin{figure}[t]
\includegraphics[width=\linewidth]{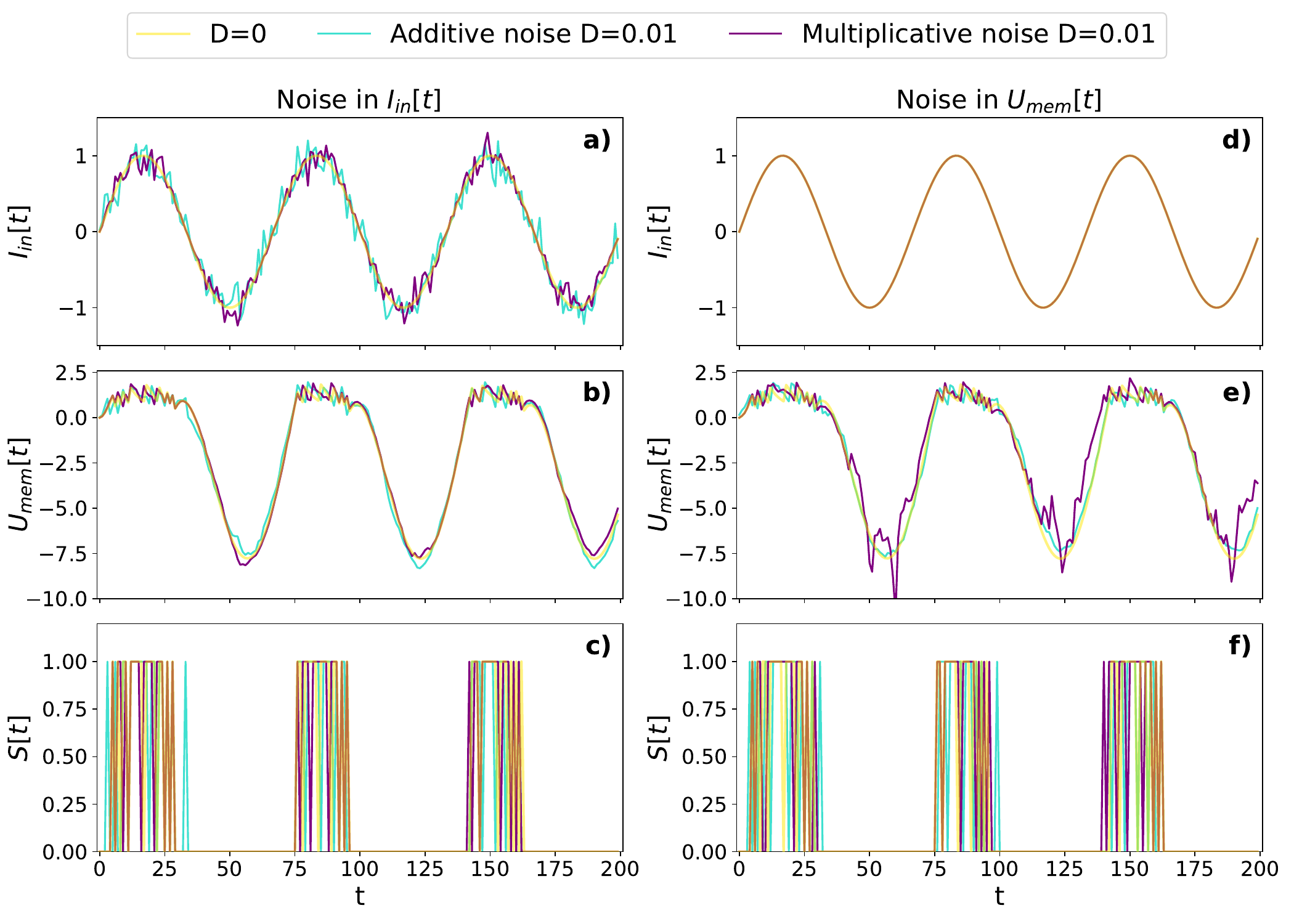}
\caption{\label{fig:single:posneg:signal} Effect of noise applied to the input current (a–c) and membrane potential (d–f). Panels show the input current (a, d), membrane potential (b, e), and output spikes (c, f) under additive noise (cyan) and multiplicative noise (purple). The noise-free case is shown in yellow. The figure corresponds to a positive input current (\ref{eq:input:posneg}).}
\end{figure}

Figure~\ref{fig:single:posneg:signal} shows $I_{\mathrm{in}}[t]$, $U_{\mathrm{mem}}[t]$, and $S[t]$ under additive or multiplicative noise applied at different stages. The figure is prepared in the same way as Fig.~\ref{fig:single:positive:signal}, but for another input current:
\begin{equation}\label{eq:input:posneg}
I_\mathrm{in}[t] = \sin[3\pi t],
\end{equation}
including several time periods and both positive and negative values (see Fig.~\ref{fig:single:posneg:signal}(a,d)).

The middle panels in Fig.~\ref{fig:single:posneg:signal}(b, e) show that for the new input signal, the membrane potential occasionally drops into negative values. However, due to the periodic nature of $I_{\mathrm{in}}$, this does not lead to complete neuron ``death.''

Comparison of noise in $I_\mathrm{in}$ shows that additive noise of a given intensity produces a constant perturbation across all input values. In contrast, multiplicative noise is most pronounced only when the input signal approaches $\pm 1$.

For noise applied to the membrane potential $U_{\mathrm{mem}}$, multiplicative noise is the most critical, as the overall range of values is much wider than in the purely positive input case (compare Fig.~\ref{fig:single:positive:signal}(e) and Fig.~\ref{fig:single:posneg:signal}(e)).

\begin{figure}[t]
\includegraphics[width=\linewidth]{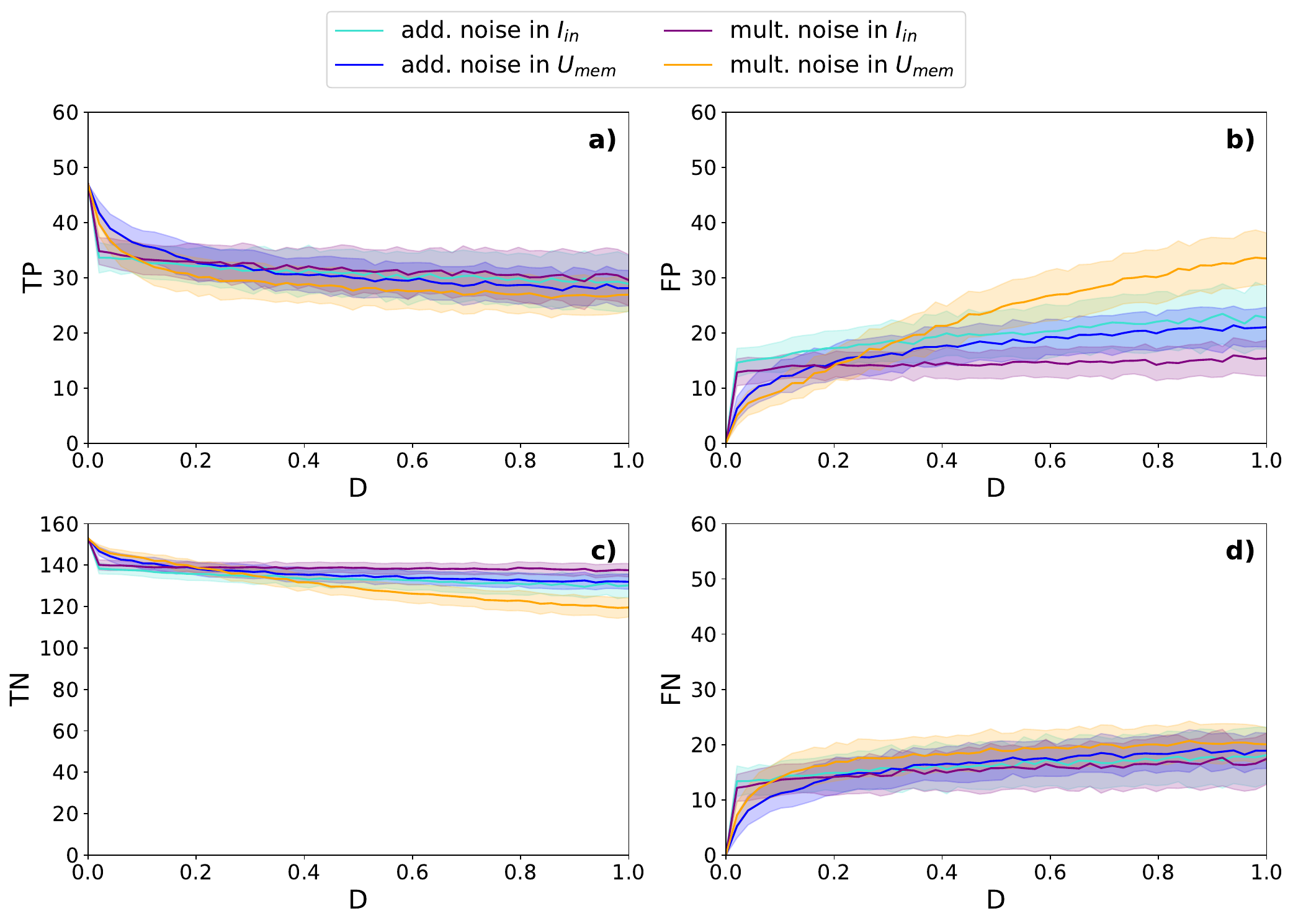}
\caption{\label{fig:single:posneg:rate} True-positive (a), false-positive (b), true-negative (c) and false-negative (d) events for a single neuron in the case additive noise influence in input current (cyan curves) or membrane potential (blue lines); and multiplicative noise influence in input current (purple lines) or membrane potential (orange lines). The curves were obtained for a positive-negative input signal (\ref{eq:input:posneg}).}
\end{figure}

Figure~\ref{fig:single:posneg:rate} illustrates how noise of varying intensities affects spike errors for the positive–negative input signal (\ref{eq:input:posneg}). The initial ranges of TP and TN differ slightly from those observed for the purely positive input (compare with Fig.~\ref{fig:single:positive:rate}). For subsequent network error analysis, the change in false positives (FP) is particularly important. Comparing panel (b) in Figs.~\ref{fig:single:positive:rate} and \ref{fig:single:posneg:rate} shows that in both cases, multiplicative noise in $I_{\mathrm{in}}[t]$ is the least critical. Additive noise in $I_{\mathrm{in}}[t]$ has a minor effect at low intensities but becomes more significant at higher $D$. As noted earlier, for positive–negative inputs, multiplicative noise in $U_{\mathrm{mem}}$ is the most critical, which is confirmed by the FP curves (orange line in Fig.~\ref{fig:single:posneg:rate}).

Interestingly, for both input types, the decrease in TP is roughly matched by the increase in FP. This suggests that if a network primarily relies on total spike count, a spiking neural network is largely resilient to noise.

\section{Trained network}\label{sec:net}
\subsection{Network topology and internal noise}\label{sec:net:topology}

To assess how the above findings generalize to a trainable network, we consider here a SNN trained on the classic MNIST handwritten digit recognition task \cite{LeCun1998}. The MNIST dataset contains 60,000 training and 10,000 test images of size $28\times28$ pixels, which determines the number of input neurons: 784. In this work, the input neurons are linear, and to introduce a temporal dimension, we assume that each input neuron’s signal is constant over a time window of $T=200$ iterations. While our future work will explore time-varying inputs, which involve additional complexities, this study focuses on stationary inputs to provide a clear understanding of filtering properties and the general effects of noise.

The number of neurons in the output layer is determined by the classification task (one neuron per class). So, 10 spiking output neurons correspond to 10 classes (digits 0–9). Since the output neurons are spiking, class membership is inferred from the neuron with the highest spike count, following a rate-coding scheme.

As for the rest of the network, we adopt a minimal topology with a single hidden layer containing 20 spiking neurons.

The network was trained using the snnTorch library with the Adam optimizer and a learning rate of $5 \cdot 10^{-4}$. The training and testing procedures are detailed in Tutorial 5 of Refs.~\cite{snntorchTutorials} and \cite{snntorch}. Training was performed over 50 epochs with a batch size of 128. The final accuracy of the trained network was approximately $94.23\%$ on the training set and $92.72\%$ on the test set.

In this section, we examine how internal noise applied to the 20 spiking neurons in the hidden layer affects the network’s final accuracy. We investigate which type of noise (additive or multiplicative) and which stage of the neuron has the strongest impact. Since the hidden layer contains multiple neurons, it is useful to distinguish between common and uncommon noise. The uncommon noise varies over time and independently across neurons (Sec.~\ref{sec:net:uncommon}), whereas common noise varies over time but is identical for all neurons (Sec.~\ref{sec:net:common}).

It is also important to note that this section presents the most representative results obtained across different trained networks and for varying numbers of neurons in the hidden layer.

\subsection{Uncommon noise}\label{sec:net:uncommon}

In this section, we examine how uncommon noise applied to the hidden layer affects the network’s accuracy. Uncommon noise varies over time and takes different values for each hidden-layer neuron at every time step.

As shown in previous sections, the input range of a neuron strongly affects how different types of noise influence its responses. Furthermore, during SNN training, we have little control over the synaptic weight matrices, and thus cannot directly regulate the input range, which can become very large and hinder analysis based on noise intensity. We also cannot control the positivity or negativity of the input signal. For this reason, we consider several strategies for pre-filtering the inputs to spiking neurons, i.e.,
\begin{equation}\label{eq:Iin_filtration}
I_{\mathrm{in},i}[t] = f(\sum\limits_{j=0}^{j<784} W_{j,i}x_j[t]).
\end{equation}
Here, the index $i$ denotes a hidden-layer neuron, and $x_j[t]$ is the signal from the previous (input) layer. The function $f(\cdot)$ represents a filtering operation. To make the analysis comparable to that performed for harmonic inputs in Sections~\ref{sec:single:positive} and \ref{sec:single:posneg}, we consider two filters with similar value ranges. The first is the hyperbolic tangent, $f(x) = \tanh(x)$, which shifts and compresses values into the range $(-1, 1)$. The second is the sigmoid, $f(x) = 1/(1+e^{-x})$, which restricts values to $(0, 1)$.

\begin{figure}[t]
\includegraphics[width=\linewidth]{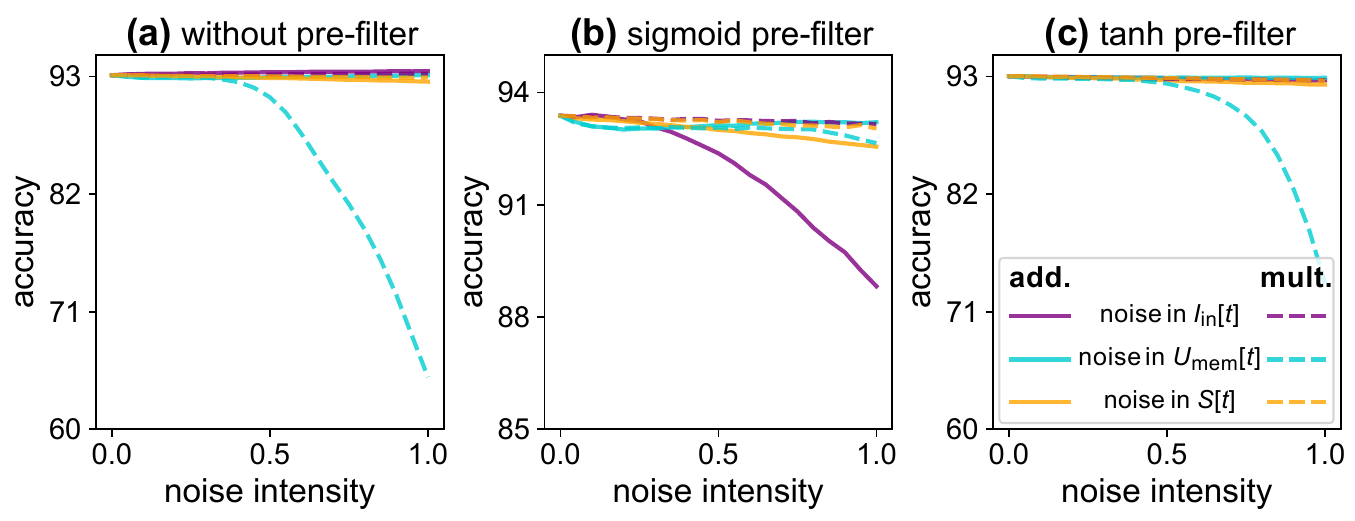}
\caption{\label{fig:snn:uncorr} Uncommon noise in trained SNN. Solid lines were prepared for additive noise while dashed lines correspond to multiplicative noise. For both noise types, the noise influence was introduced into input current (purple), membrane potential (cyan) and neurons' output (orange). The results were prepared without additional filtration (a), with sigmoid pre-filter (b) and tanh pre-filter (c). }
\end{figure}

Figure~\ref{fig:snn:uncorr} shows the network accuracy as a function of noise intensity. Different colors indicate noise applied to different parts of the spiking neuron, analogous to the single-neuron analysis in previous sections: the input current $I_{\mathrm{in}}[t]$, the membrane potential $U_{\mathrm{mem}}[t]$, and, additionally, the output spiking signal $S[t]$, which is meaningful only in the context of a network. Solid lines represent additive noise, while dashed lines correspond to multiplicative noise. Each point of curve is averaged over 10 repetitions for each input image.

The panels in Fig.~\ref{fig:snn:uncorr} show results for the SNN without input filtering in the hidden layer (a), with a sigmoid pre-filter (b), and with a hyperbolic tangent pre-filter (c). Comparing panels (a) and (c), the effect of noise is roughly the same for both networks, likely due to the presence of negative inputs and the resulting neuron death. For a single neuron, multiplicative noise in the membrane potential was the most critical; the network shows a similar pattern (see the cyan dashed line in Fig.~\ref{fig:snn:uncorr}(a, c)). The other curves across these panels are nearly identical. At low noise intensities, additive noise in the membrane potential (solid cyan lines) is slightly more critical, whereas at high intensities, additive noise in the output spiking signal $S[t]$ (solid orange lines) dominates. Importantly, overall accuracy varies by only about 1\% even at $D=1$.

When a sigmoid pre-filter is applied (Fig.~\ref{fig:snn:uncorr}(b)), the network becomes significantly more resistant to noise. The most critical case is additive noise in the input current (solid purple line), yet even here accuracy drops by only about 5\% at $D = 1$. For the other types of noise, accuracy declines by around 1\% for additive noise in the output spikes (solid orange line) and multiplicative noise in the membrane potential (dashed cyan line). All other noise types have negligible impact, including additive noise in the membrane potential, which was more critical for a single neuron (see Fig.~\ref{fig:single:positive:rate}). This noise resistance may arise because, as noise intensity increases, false-positive spikes occur across all noisy neurons, but in a rate-coded network, the neuron with the highest spike count remains largely unchanged.


\subsection{Common noise}\label{sec:net:common}

\begin{figure}[t]
\includegraphics[width=\linewidth]{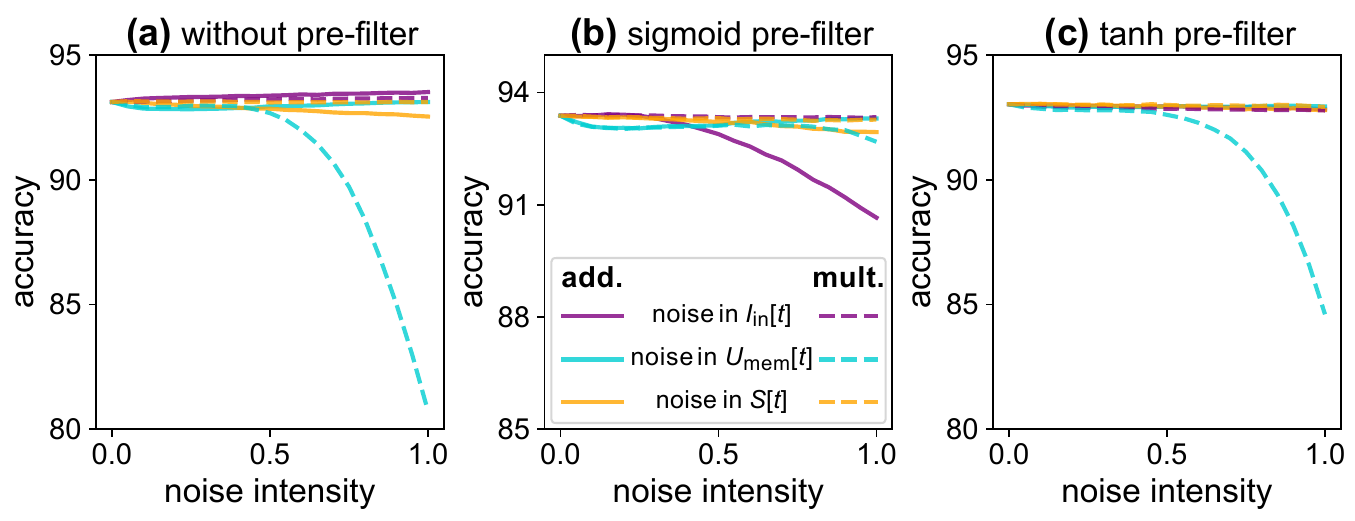}
\caption{\label{fig:snn:corr} Common noise in trained SNN. Solid lines were prepared for additive noise while dashed lines correspond to multiplicative noise. For both noise types, the noise influence was introduced into input current (purple), membrane potential (cyan) and neurons' output (orange). The results were prepared without additional filtration (a), with sigmoid pre-filter (b) and tanh pre-filter (c). }
\end{figure}

Similarly, the effect of common noise on the trained SNN was investigated. These results are shown in Fig.~\ref{fig:snn:corr}. Common noise varies over time but is identical across all 20 hidden-layer neurons.

Comparing Fig.~\ref{fig:snn:uncorr} (unshared noise) with Fig.~\ref{fig:snn:corr} (shared noise), the shape and relative arrangement of the curves are very similar. The main difference is that common noise causes a much smaller decrease in accuracy compared to uncommon noise. It is also notable that, with a pre-filter, common noise in the output spikes $S[t]$ is less critical than in the uncommon case. Similar trends were observed across other trained networks, while Figs.~\ref{fig:snn:uncorr} and \ref{fig:snn:corr} show the most representative examples.

\section{Conclusion}\label{sec:conclu}

This work investigated the effects of additive and multiplicative noise on a single LIF neuron and a trained SNN. Noise was introduced at different stages of the neurons: the input current $I_{\mathrm{in}}[t]$, the membrane potential $U_{\mathrm{mem}}[t]$, and the output spikes $S[t]$ of noisy neurons in the network. We found that multiplicative noise in the membrane potential is the most critical factor for network performance, causing the largest drop in accuracy. This occurs because a reduction in membrane potential can drive a neuron toward ``death'' and large negative values, and the multiplicative nature of the noise amplifies its overall effect. One effective strategy to mitigate this impact is to apply a pre-filter at the input of the noisy neuron. The sigmoid filter performed best, shifting the input range into the strictly positive domain. In this case, the most critical noise becomes additive noise in the input current of the noisy neurons $I_{\mathrm{in}}[t]$. All other combinations of noise stage and type reduced network accuracy by no more than 1\%, even at a relatively high noise intensity $D=1$, comparable to the input signal range.

The study also examined two types of noise with respect to their impact across neuron populations: uncommon and common noise. Qualitatively, the effects were similar, but quantitatively, common noise caused a much smaller reduction in accuracy than uncommon noise. This indicates that SNNs are inherently robust to common noise.

In summary, this work identified the noise types most critical to spiking neural networks and proposed strategies to mitigate their impact on network accuracy.

\begin{acknowledgments}
This work was supported by the Russian Science Foundation (project No. 25-72-10055). 
\end{acknowledgments}

\section*{Data Availability Statement}
The data that support the findings of this study are available from the corresponding author upon reasonable request.

\bibliography{bibliography}

\end{document}